\documentclass[10pt,twocolumn,letterpaper]{article}

\usepackage{iccv}
\usepackage{times}
\usepackage{epsfig}
\usepackage{graphicx}
\usepackage{amsmath}
\usepackage{amssymb}
\usepackage[accsupp]{axessibility}  % Improves PDF readability for those with disabilities.

\usepackage[breaklinks=true,bookmarks=false]{hyperref}

\iccvfinalcopy % *** Uncomment this line for the final submission

\def\iccvPaperID{****} % *** Enter the ICCV Paper ID here

\ificcvfinal\fi

\begin{document}

%%%%%%%%% TITLE
\title{Weakly-Supervised Action Localization by Hierarchically-structured Latent Attention Modeling}

\author{
Guiqin Wang$^1$\footnote[2]{Work is done during the internship at ACS Lab, Huawei Technologies.}
\quad Peng Zhao$^1$
\quad Cong Zhao$^{3,4}$
\quad Shusen Yang$^{3,4}$
\quad Jie Cheng$^2$ 
\quad Luziwei Leng$^2$\\
\quad Jianxing Liao$^2$
\quad Qinghai Guo$^2$\footnote[1]{Corresponding author (guoqinghai@huawei.com).}\\
$^1$ School of Computer Science and Technology, Xi'an Jiaotong University\\
\quad $^2$ ACS Lab, Huawei Technologies \\
\quad $^3$ School of Mathematics and Statistics, Xi'an Jiaotong University\\
\quad $^4$ National Engineering Laboratory for Big Data Analytics, Xi'an Jiaotong University
}

\maketitle
\ificcvfinal\thispagestyle{empty}\fi
{
  \renewcommand{\thefootnote}%
    {\fnsymbol{footnote}}
  \footnotetext[2]{Work is done during the internship at ACS Lab, Huawei Technologies.}
  \footnotetext[1]{Corresponding author (guoqinghai@huawei.com).}
}

%%%%%%%%% ABSTRACT
%%%%%%%%% ABSTRACT
\begin{abstract}
Weakly-supervised action localization aims to recognize and localize action instancese in untrimmed videos with only video-level labels. Most existing models rely on multiple instance learning(MIL), where the predictions of unlabeled instances are supervised by classifying labeled bags. The MIL-based methods are relatively well studied with cogent performance achieved on classification but not on localization. Generally, they locate temporal regions by the video-level classification but overlook the temporal variations of feature semantics. To address this problem, we propose a novel attention-based hierarchically-structured latent model to learn the temporal variations of feature semantics. Specifically, our model entails two components, the first is an unsupervised change-points detection module that detects change-points by learning the latent representations of video features in a temporal hierarchy based on their rates of change, and the second is an attention-based classification model that selects the change-points of the foreground as the boundaries. To evaluate the effectiveness of our model, we conduct extensive experiments on two benchmark datasets, THUMOS-14 and ActivityNet-v1.3. The experiments show that our method outperforms current state-of-the-art methods, and even achieves comparable performance with fully-supervised methods. 
\end{abstract}

%%%%%%%%% BODY TEXT
\section{Introduction}

\label{sec:intro}

Action localization is one of the most challenging tasks in video analytics and understanding~\cite{islam2021hybrid, yang2021uncertainty, zhai2020two, ji2021weakly}. The goal is to predict the accurate start and end time stamps of different human actions. Owing to its wide application (\textit{e.g.}, surveillance~\cite{vishwakarma2013survey,yang2022massive}, video summarization~\cite{ma2005generic}, highlight detection~\cite{hong2020mini}), action localization has drawn lots of attention in the community. To tackle this problem, many methods try to solve it in a fully-supervised manner~\cite{chen2019relation, zeng2019graph, zeng2021graph}, but they rely on massive time-consuming annotations. To alleviate this issue, researchers pay more attention to weakly-supervised action localization (WSAL)~\cite{zeng2019breaking,islam2021hybrid,ji2021weakly,yang2021uncertainty,zhai2020two,lee2021weakly,yang2022uncertainty,He_2022_CVPR,li2022exploring}, which explores a more efficient learning strategy with only video-level categorical labels. Nevertheless, recent works~\cite{lee2021weakly,He_2022_CVPR,zhai2020two,li2022exploring} mostly rely on the multiple instance learning (MIL) framework~\cite{zhou2004multi}: obtaining a video-level prediction via aggregation and optimization under the video-level supervision.
\begin{figure}[ht]
  \centering
  \includegraphics[width=0.45\textwidth]{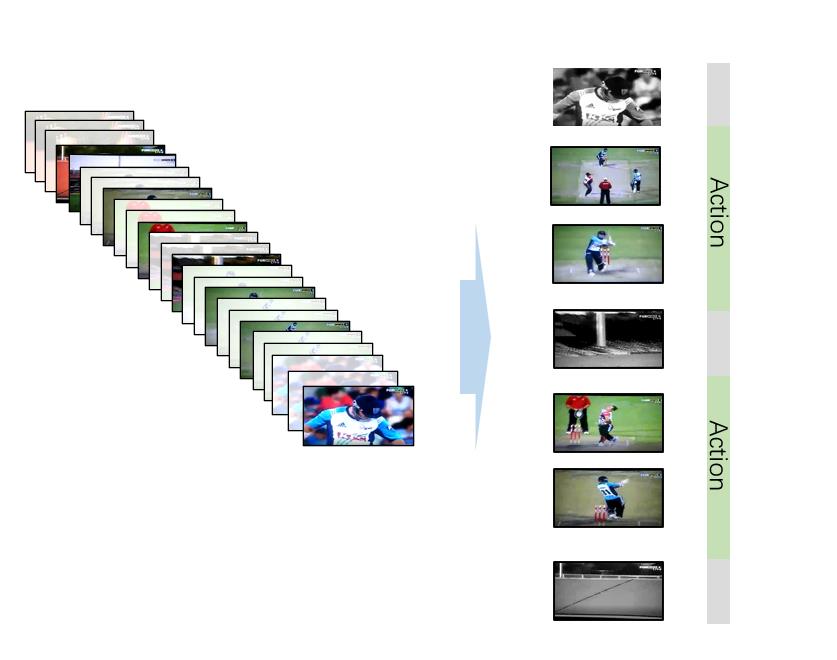}
  \caption{Visualization of the change-point detection component and the co-occurrence component(attention-based classification module) decoupled by our method from a snippet representation. The change-point component helps to detect change-points of temporal variations, which include the change-points of foreground in a video(\textit{i.e.} the highlighted frames in the left part). Collaborating with the attention module, the points of the foreground are chosen as boundaries of action(\textit{i.e.} the highlighted frames in the right part).}
  \label{fig:1}
\end{figure}
While significant improvement has been made in prior MIL-based work, there is still a huge performance gap between the weakly-supervised and fully-supervised settings. In consideration of this issue, diverse solutions have been proposed in the literature. For instance, ~\cite{zeng2019breaking,islam2021hybrid,ji2021weakly} try to erase the most discriminative parts for learning action completeness, \cite{yang2021uncertainty,zhai2020two, he2022asm} learn with pseudo labels generated by manual thresholds and iterative refinement, and~\cite{qu2021acm,shi2020weakly,zhao2022temporal} formulate the WSAL problem as a video recognition problem and introduce an attention mechanism to construct video-level features, then apply an action classifier to recognize videos.

All the above approaches largely rely on the video-level classification model, which aims at learning the effective classification functions
to identify action instances from bags of action instances and
non-action frames, but overlook the significance of feature variations. In fact, features usually contain intense semantic information~\cite{chen2019relation,quader2020weight}, mainly stemming from the temporal and spatial variations in video actions. The variation of features is useful for correcting the wrong action region and adjusting the imprecise boundary of the temporal proposal. Existing solutions often neglect such semantics and thus largely suffer from deviated action boundaries and inaccurate detection.
%neither do they consider the temporal variations of such semantics around the action boundaries in a (latent) state space.

To better learn the semantics in a given video sample, the model should be able to encode the temporal variations of different time factors. Intuitively, these variations in different timescales disentangle different video fragments, and detection on such variations automatically leads to the detection of change-points, which provide the candidates for action boundaries.
% To better utilize the feature semantics in a WSAL task, one needs to develop a model to learn the temporal variations and detect the  
% one needs to develop a representation model to learn the spatial and temporal features precisely as a first step. Recent representation learning work show that incorporation of the hierarchy into variational latent models lead to extremely expressive representations of spatiotemporal data~\cite{zakharov2021variational,saxena2021clockwork,vahdat2020nvae}. 
% and the variations can be represented by a hierarchical temporal structure where hierarchies refer to timescales

Derived from the above idea, we propose a novel Attention-based Hierarchically-structured Latent Model (AHLM) to model the spatial and temporal features for WSAL task. Specifically, we detect the action boundaries as the change-points of a generative model, where those change-points are determined at the time points with inaccurate generation. Such generative model, is trained by learning the hierarchical representations of the feature semantics in the latent space based on the video inputs. By using an attention-based classification model to select the change-points of the foreground, AHLM localizes the exact action boundaries, see Figure~\ref{fig:1} for an illustration. %, these change-points are essentially equivalent to the action boundaries. %By a collaboration with the attention-based classification model, we successfully obtain the required video clips and their categorical labels. 
%by leveraging this variation of action videos, our model is trained to represent variation of action videos, that relies on a change detection mechanism to impose a nested temporal hierarchy on its latent representations.~\cite{zakharov2021variational,saxena2021clockwork}

To our best knowledge, we are the first to consider the temporal variation of feature semantics and study the change-point detection mechanism in WSAL. We design an AHLM that prominently boosts WSAL performance. Our main contributions are summarized as follows:

\begin{itemize}
  
\item
To leverage the temporal variations of feature semantics for WSAL, we propose a hierarchically-structured generative latent model that explores spatiotemporal representations and leverages the temporal feature semantics to detect the change-points of videos.

\item
We build a new framework, AHLM, which firstly proposes the use of an unsupervised change-points detector in a latent space to complement weakly supervised learning, with a novel hierarchical generative model-baed change-point detector for complex datasets. 

\item
Based on extensive experiments, we demonstrate that, on two popular action detection datasets, our novel 
AHLM provides considerable performance gains. On THUMOS14 especially, our method achieves an average mAP of $47.2\%$ when IOU is from $0.1$ to $0.7$, which is the new state-of-the-art (SotA). On ActivityNet v1.3, our method also achieves the new SotA, with an average mAP of $25.9\%$ when IOU is from $0.5$ to $0.95$. 
\end{itemize}

% The remainder of the paper is organized as follows: in Section~\ref{sec:Related work}, we discuss related studies on weakly-supervised action localization, unsupervised action analysis, and hierarchical generative model, respectively. In Section~\ref{sec:Method}, we present the details of our model's overview and theoretical basis. In Section~\ref{sec:Experiment}, we report the implementation details, evaluation results, and ablation study. In Section~\ref{sec:Conclusion}, we conclude the paper.

\section{Related work}
\label{sec:Related work}

\subsection{Weakly-supervised Action Localization}
Due to the precise annotation of each action instance in fully-supervised, the Weakly-supervised Action Localization(WSAL) is proposed to reduce the expensive annotation costs. During training, the WSAL methods~\cite{chen2019relation,quader2020weight,yang2021uncertainty,zhai2020two} require only video-level categorical labels. These methods can be grouped into two categories, namely top-down and bottom-up methods. In the top-down pipeline, the video-level classification model is learned first, and then frames with high classification activation values are selected as action locations.~\cite{paul2018w} and~\cite{narayan20193c} forced foreground features from the same class to be similar, otherwise dissimilar. Unlike the top-down scheme, the bottom-up methods directly produce the attention for each frame from data, and train a classification model with the features weighted by attention. Based on this paradigm,~\cite{nguyen2018weakly} further added a regularization term to encourage the sparsity of action.~\cite{yuan2019marginalized} proposed a method to suppress dominance of the most salient action frames and retrieve less salient ones.~\cite{shi2020weakly} proposed a model to learn the class-agnostic frame-wise probability conditioned on the frame attention using conditional Variational Auto-Encoder (VAE). Nevertheless, all of the aforementioned methods overlook the significance of temporal variations with respect to features. Unlike these methods, we focus on modeling the temporal variations of the feature semantics, and utilize an unsupervised change-point detection method to localize the action boundaries.

\subsection{Unsupervised Action Analysis}

Unsupervised learning targets learning effective feature representations from unlabeled data. \cite{sarfraz2021temporally} proposed a temporally-weighted hierarchical clustering algorithm to represent actions in the video.~\cite{du2022fast} estimated the similarities across smoothed frames through the difference of actions and external discrepancy across actions. These methods, however, mainly focused on the variation of the frame(\textit{e.g.}, the similarity of frame, the temporal variation
of frame). Similarly,~\cite{aakur2019perceptual} utilized spatial-temporal dynamics of events to learn the visual structure of events in the latent space. However,~\cite{aakur2019perceptual} mainly modeled simple datasets(\textit{i.e.}, Breakfast Actions, 50 Salads, and INRIA Instructional Videos), as the proposed method lacks the capability to represent heterogeneous information in an entangled latent space. Unlike the above methods, we utilize the hierarchically-structured VAE and subjective timescaled transition model to learn spatiotemporal semantics on the multi-scaled latent space, which expands the method’s applicability to more complex datasets(\textit{i.e.}, Thumos14 and Activitynet1.3).
 
\subsection{Hierarchical Generative Model}
The hierarchical generative model has experienced a fast development in recent years~\cite{saxena2021clockwork,vahdat2020nvae, zakharov2021variational}, due to the fact that incorporating hierarchy into latent models improves the expressiveness of spatiotemporal representation. \cite{vahdat2020nvae} implemented a stable fully-convolutional and hierarchical VAE with the use of separate deterministic bottom-up and top-down information channels. \cite{saxena2021clockwork} was designed to model temporal data using a hierarchy of latent variables that update over fixed time intervals. Nevertheless, these methods were designed to focus on temporal variations while barely considering temporal semantics. Quite recently, VPR~\cite{zakharov2021variational} proposed a subjective timescale-based hierarchical structure to model the different temporal dynamics, however, they handled only local information of simple data(\textit{e.g.}, bouncing balls). Consecutive videos, when actions change, have more complex global semantic information leading to detecting more precise boundaries. In our work, we propose to simultaneously model the temporal dynamics and the varied distributions of global semantics which enables the hierarchical generative model to the WSAL tasks. 

% \cite{chung2019hierarchical} proposed Hierarchical Multiscale RNNs and showed that a modified discrete-space LSTM can discover multiscale structure in data in an unsupervised manner. 
%-------------------------------------------------------------------------

\section{Method}
\label{sec:Method}

% In this section, we first define the formulation of WSAL, then present our attention-based  hierarchically-structured latent model (AHML) in detail, and thereafter, introduce the training and inference details. The overview architecture of our AHML is illustrated in~\ref{fig:2}.
\begin{figure*}
  \centering
  \includegraphics[width=0.95\textwidth]{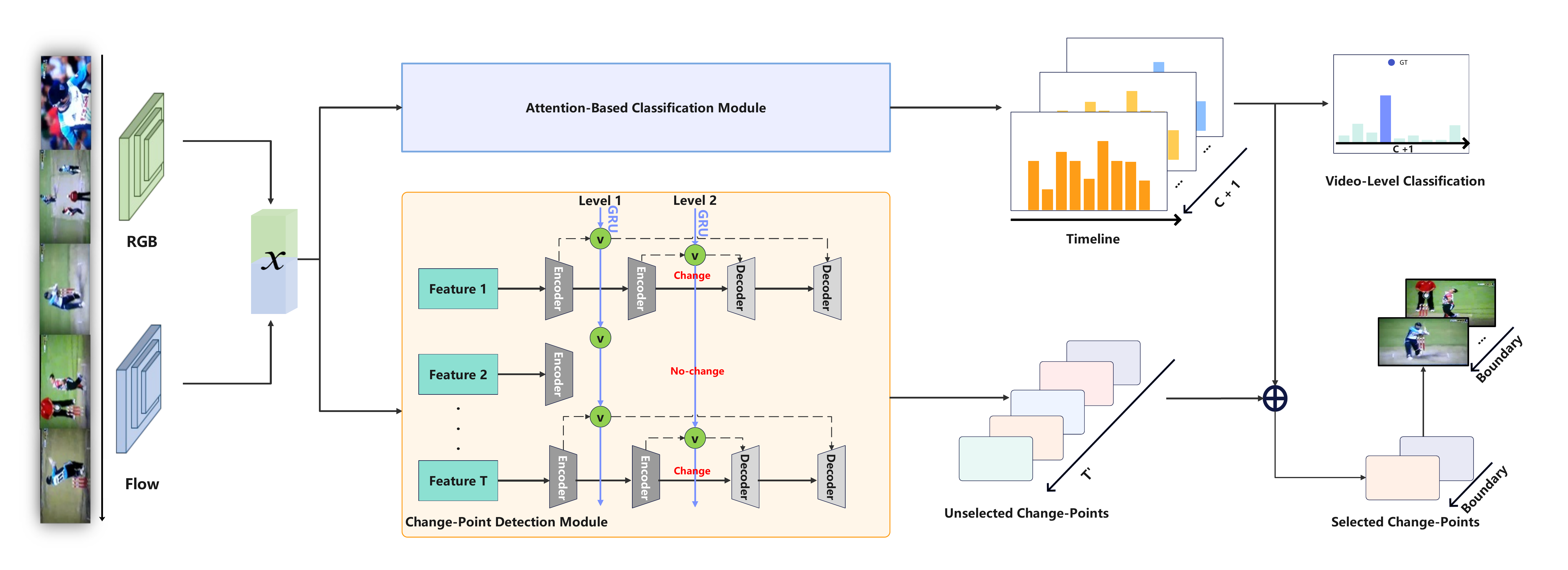}
  \caption{The overall pipeline of the proposed framework. It consists of three parts: Feature Embedding, Change-point Detection Module(DFC), and Attention-based classification Module(EFC). First, the feature embedding stage extracts original snippet features through the I3D network. Subsequently, the DFC is trained to represent spatiotemporal information and change points of feature semantics, supervised by feature distribution and feature reconstruction. Meanwhile, the EFC is trained to distinguish foreground, background and context. For inference, the DFC produces change-points and the EFC selects the change-points of the foreground as boundaries.}
  \label{fig:2}
\end{figure*}
% \subsection{Problem Formulation}
Suppose we have a set of training videos and the corresponding video-level labels. Specifically, let us denote for an untrimmed training video, its ground-truth label as $y \in \mathbb{R}^C$, where $C$ is the number of action categories. Note that $y$ could be a multi-hot vector if more than one action is presented in the video and is normalized with the $l_1-$normalization. The goal of temporal action localization is to generate a set of action segments $ S =\left\{ (s_i, e_i, c_i, q_i) \right\}^{I}_{i=1}$ for a testing video with the number of $I$ segments, where $s_i$, $e_i$ are the start and end time point of the $i$-th segment and $c_i$, $q_i$ are the corresponding class prediction and confidence score.

Our method follows the bottom-up pipeline for WSAL, where we detect change-points $ P =(p_i)_{i \in T^{\prime}}$ directly from data. Here $p_i$ is the change-point frame of the video, which is detected based on the transitions of the observable features, and $T^{\prime} $ is the set of times when a changed-point occurs. Then we leverage an attention model to optimize change-points of the video to obtain the refined boundaries. 
% Before discussing the details of our method, we examine the action localization problem from the beginning.

\subsection{Framework Overview}

In the localization problem, the target is to predict the boundaries of the action instance, which is essentially equivalent to solving the boundary problem of semantic representations for those instances. To this end, our proposed Attention-based Hierarchically-structured Latent Model(AHLM) divides this problem into two different aspects for boundaries localization, the detection of feature semantic change-points(DFC), and the extraction of foreground change-points(EFC). The second term \textbf{EFC} is optimized by discriminative capacity for action classification, which is the main optimization target in previous works~\cite{lee2020background, qu2021acm, he2022asm}. In contrast, the first term \textbf{DFC} forces the representation of spatial features to be accurately predicted from the temporal changes, which requires the capability of feature disentanglement. In particular, to learn disentangled spatiotemporal representations, we exploit an action-based hierarchical VAE model to encode the input as hierarchical latent spaces, then construct GRU-based transition models 
which learn to predict feature changes in the latent spaces optimally.

As Figure~\ref{fig:2} shows, AHLM is a two-branch network, including a change-point detection module and an attention-based classification branch. Given the concatenated features $\mathit{X} \in \mathbb{R}^{T \times 2D}$ with $T$ snippets for a video, the change-point detection module predicts the frame-level change-points  $ P = [p_1, p_2, ..., p_{T\prime}] $, conveying the class-agnostic change-point boundaries. In the attention-based classification model, input features $X$ are used to predict an attention-based snippet-level class activation map $M$ by a classification head with background class~\cite{qu2021acm, shi2020weakly}, which learns the frame attention by optimizing the video-level recognition task. Note that, $M=\left\{m_{c,i}\right\}^{C+1,T} \in \mathbb{R}^{C+1,T}$, where $m_{c,i}$ is the activation of $c$-th class for $i$-th snippet (the $C+1$-th class means the background category). By suppressing the background change-points $P$ via the attention-based class activation map $M$, we can get foreground-based change-points as boundaries of action instances.

With predictions generated as the description above, we further explore the hierarchical-structured detection model (\textit{i.e.} Change-point Detection Module) and attention-based classification module, to facilitate the learning of action classification and boundaries localization.

\subsection{Change-point Detection Module}
\begin{figure}
  \includegraphics[width=0.45\textwidth]{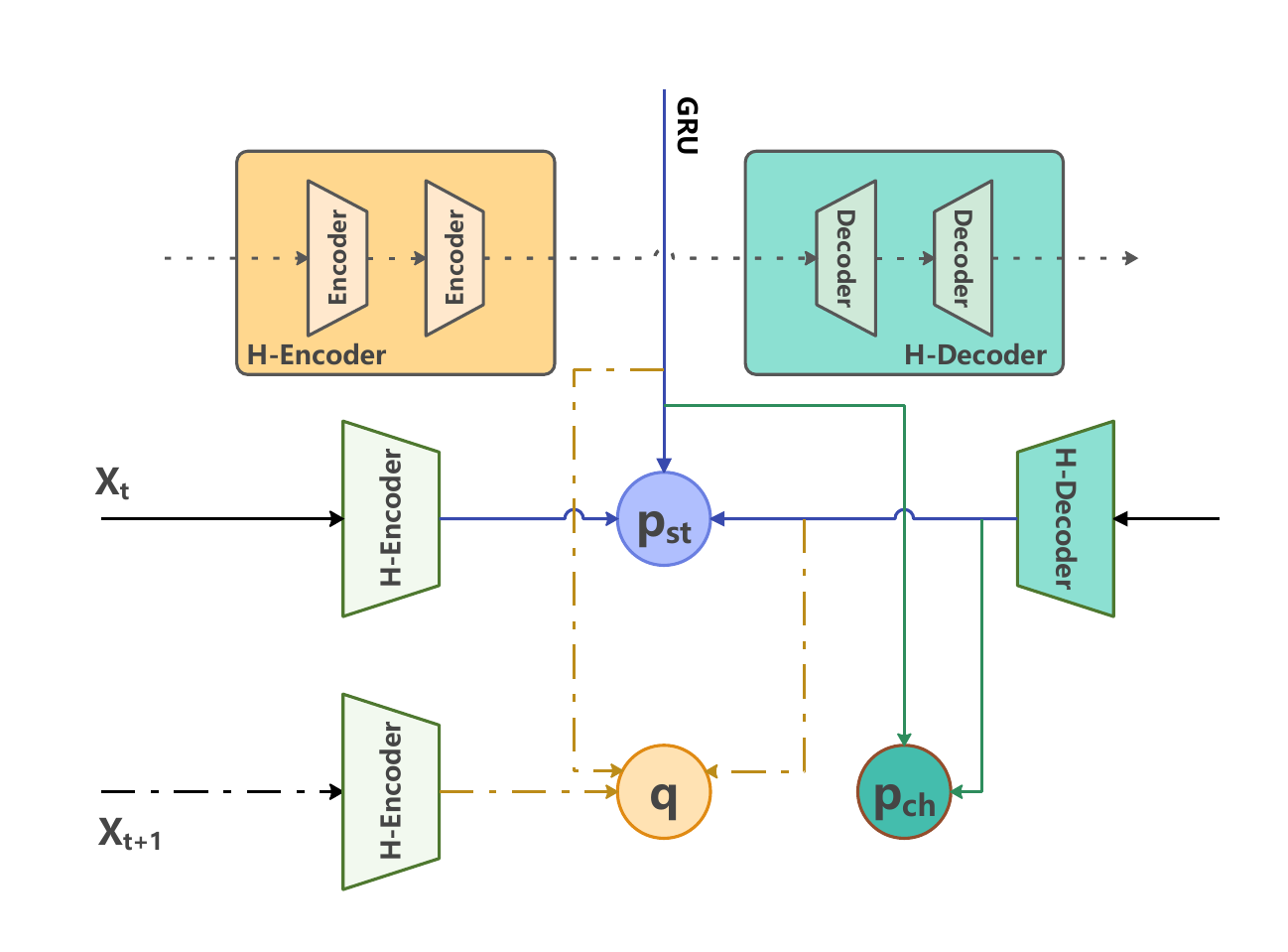}
  \caption{Change-point Detection Module. The black dotted lines indicate the architecture of the Change-point Detection Module. The colored lines indicate the mechanism of the Change-point Detection Module, where dotted lines indicate the next state $x_{t+1}$, and solid lines indicate the current state $x_t$. The $p_{st}$, $p_{ch}$ and $q$ represent the distribution of, current state $x_t$, predicted result, and next state $x_{t+1}$ in the latent space respectively.}
  \label{fig:3}
\end{figure}
The cornerstone of supervised learning is to fully leverage the given annotations, especially for the weakly-supervised learning that has limited information. Previous works mainly develop their framework on MIL-paradigm for video-level learning, ignoring the temporal variation of features. Our framework is constructed under the following cognitive phenomenon: \textit{if one can build a perfect world model, then the change-points will occur when this model does not achieve an ideal prediction along the temporal domain.} 
Indeed, a perfect prediction should clearly reﬂect the information at both video-level and feature-level. Such feature-level learning, mainly unsupervised, relies on an effective expression of spatiotemporal information. Recent hierarchical generative models~\cite{saxena2021clockwork,zakharov2021variational} show strong capability regarding this aspect, which make it possible to construct an event boundary detection module based on the predictions.
% this aspect, while the resulting temporal feature disentanglement automatically induces the event boundary detection. 
Inspired by such observations, we develop our change-point detection module based on a hierarchical generative model. We build a $2$-level generative model through a variational autoencoder(VAE) structure, combined with a transition model to learn the temporal variations of the video. The first level aims to learn the latent representation for each time point, and the second level further projects the encoded information to higher latent space when observed a change-point based on DFC. Figure~\ref{fig:3} indicates the mechanism and architecture of the DFC. 
 % Our module calculates the KL-divergence between the current state and next state on the latent space to detect change point.
 % Each block in level $n$ ($n=1,2$) consists of three deterministic variables $(x_t^n, u_t^n, d_t^n)$ and a random variable $v_t^n$, corresponding to three channels of communication between the blocks: bottom-up (encoding), top-down (decoding), and temporal (transitioning), and learned latent representation, respectively. $v_t^n$ is parameterized by $(x_t^n, u_t^n, d_t^n)$ in the same level and used to sample $u_t^{n-1}$ in the lower level and $d_{t+1}^n$ in the next time point.

%The hierarchical action-based generative model consists of computational blocks that are stacked together in a hierarchy with several levels. Figure~\ref{fig:3} shows the insides of a single block with key variables and channels of information flow. Each block in level $n$ consists of three deterministic variables $(x_t^n, o_t^n, d_t^n)$ that represent the three channels of communication between the blocks: bottom-up (encoding), top-down (decoding), and temporal (transitioning), respectively. These variables are used to parameterize a random variable $s_t^n$ that contains the learned representations in a given hierarchical level.

% For the architecture of the DFC, 
As Figure~\ref{fig:3} shows, we use H-Encoder(two $1024$-d fully-connected networks) to encode the feature $x$ through $f = f_{H-enc}(x)$, and use H-Decoder(two $1024$-d fully-connected networks) to decode the feature $u$ based on the encoder output $f$, namely, $u=f_{H-dec}(f)$. We use a recurrent GRU model~\cite{cho2014learning}($256$-d) to learn the temporal transition through $d_{t+1}=f_{tran}(v_t,d_t)$, where $t$ indicates temporal dimension, and $v_t$ is the latent random variable conditioned on the deterministic variables $f_t$, $u_t$ and $d_t$. For a given sequence of input observations $\left\{x_1, x_2, ..., x_T, x_{T+1} \right\}$, modeled by the latent variables $v_{1:T}^{1,2}$, where superscripts $1$, $2$ indicate different levels and subscript $1:T$ indicates time sequence, the generative model can be written as the following factorized distribution:
% between the blocks is a crucial component of the system. Top-down decoding from level $n + 1$ to $n$ is realised by passing the latest context $o^{n+1}$ and sample from $s^{n+1}$ through a neural network to retrieve $o^n=f_{dec}(o^{n+1},s^{n+1})$. Temporal transitioning is implemented with the use of a recurrent GRU model~\cite{cho2014learning}, such that $d_{t+1}=f_{tran}(s_t,d_t)$. Finally, bottom-up encoding of new observations iterative computes layer-wise observation variables, $x^{n+1}=f_{enc}(x^n)$. $o_t^n$ and $d_n^t$ are therefore used to represent $s_t^{>n}$ and $s_t^{<n}$ respectively. This means that for a sequence of input feature $\left\{x_1, x_2, ..., x_T \right\}$, the hierarchical action-based generative model can be written as the following factorised distribution:
% 
% \begin{equation}
%     p (o_{1:T}, s_{1:T}^{1,2}) = [\prod_{t=1}^T p(o_{T_1}|s_{T_1}^{1,2})][\prod_{n=1}^2\prod_{t=1}^T p(s_t^n|s_{<t}^n,s_t^{>n})],
% \end{equation}
% 
\begin{align}\label{eq:gen}
p(x_{1:T}&, v_{1:T}^1, v_{\tau_1:T_2}^2) = [\prod_{t=1}^T p(x_{t} | v_t^1, v_t^2)]\cdot \\ \notag
&[\prod_{t=1}^T p(v_t^1 | v_{1:t-1}^1, v_t^2)]\cdot 
[\prod_{t=\tau_1}^{T_2} p(v_t^2 | v_{t'<t}^2)],
\end{align}
where $T_2$ denotes the number of change-points detected by level $2$, and we define the distribution of the initial state of $v_1^{2}$ as $p(v_1^{2}) = \mathcal{N}(0,1)$ a Gaussian prior. Note that $p(v_t^1|v_{1:t-1}^1, v_t^2)$ is a prior distribution of the latent state $v_t^1$ conditioned on all the past states $v_{1:t-1}^1$ in level $1$ and the possible past upper level state $v_{t}^2$, and $p(v_t^2|v_{t'<t}^2)$ is the prior distribution of the latent state $v_t^2$ conditioned on all the past states $v_{t'<t}^2$, which are all the possible states in the same level $2$ before time $t$.

% Since it is impossible to tract the true posterior $p(v_{1:T}^1, v_{1:T'}^2 | x_{1:T})$ given the observations $x_{1:T}$, we leverage to approximate it using an approximate posterior distribution $q$, defined by 
% \begin{align}\label{eq:posterior}
%     q(v_{1:T}^1, v_{1:T'}^2 | x_{1:T}) =& \prod_{t=1}^T q_{\phi}(v_t^1 | x_t^1, v_{1:t-1}^1, v_{t'<t}^2)\cdot \notag \\
%     & \prod_{t=1}^{T'} q_{\phi}(v_t^2 | x_t^2, v_{t'<t}^2),
% \end{align}
% where $\phi$ represents the parameters of the posterior model and $x_t^n$ is the level-wise encoding of $x_t$.
Follow the general VAE~\cite{kingma2013auto}, 
% we approximate the intractable true posterior $p_\theta(v_t^n|x_t, v_t^{>n}, v_{<t}^n)$ using a posterior distribution $q_\phi(v_t^n|x_t,v_t^{>n},v_{<t}^n)$, and 
we define the corresponding variational evidence lower bound(ELBO) is as:
\begin{equation}\label{eq:elbo}
    \begin{aligned}
        & \mathcal{L}_{ELBO} = \sum_{t=1}^T \mathbb{E}_{q(v_t^{1,2})}[\log p(x_t|v_t^{1,2}]- \sum_{n=1}^2\sum_{t=1}^T \\ &\mathbb{E}_{q(v_t^{>n},v_{<t}^n)}[D_{KL}(q_\phi(v_t^n|x_t,v_t^{>n},v_{<t}^n)||p_\theta(v_t^n|v_t^{>n},v_{<t}^n))],
    \end{aligned}
\end{equation}
where $p_\theta$ is the prior model with $\theta$ representing the parameters defined by $x,u$ and $d$, $q_\phi$ is the posterior model and $v_t^{>n}$ represents the latent states at time $t$ in the level higher than $n$. 

The first term in Equ.~\eqref{eq:elbo} represents the likelihood of the reconstructed $x_t$ given the latent variables $v_t^{1,2}$, which measures the reconstruction loss. The second term is the KL-divergence of the prior distribution $p(v)$ and the posterior distribution $q(v)$. The loss function of DFC defines by
\begin{equation}\label{eq:DFC}
\mathcal{L}_{DFC} = -\mathcal{L}_{ELBO}.
\end{equation}

Our model is trained in a way that the second level updates the latent state $v_t^2$ in a subjective time scale, while the determined time points correspond to the changes in the observable features over time. 
% \textcolor{blue}{GRU reset; adaptive parameters of $\beta$; introduction - temporal representation, inspired by cortex mechanism; attention - rewrite our contribution }

The key component of the DFC relies on a Bayesian inference mechanism under the static assumption on the level $2$, and the changes are detected when the updated posterior violates such an assumption. Specifically, as Figure~\ref{fig:3} shows, given the current feature inputs from the H-Encoder $f_t$ and H-Decoder $u_t$, we construct the static assumption and change assumption respectively (note we omit the level index for better readability in the following since we only consider the detection in level $2$). Under the static assumption, the prior is calculated as 
\begin{equation}\label{eq:p_st}
    p_{st}= p_\theta(v_{t+1} | f_t, d_t, u_t),
    % \simeq q_\phi (v_{t} | f_t, d_t, u_t),
    % p_{st}= p(v_{t+1}^2 | x_t^2, v_{1:t-1}^2) \simeq q_\phi (v_{t}^2 | x_t^2, v_{1:t-1}^2),
\end{equation} 
while under the change assumption, the prior is calculated as
\begin{equation}\label{eq:p_ch}
    p_{ch} = p_{\theta}(v_{t+1} |f_t, d_{t+1}, u_t), 
    % \simeq q_\phi (v_{t+1} | f_{t+1}, d_{t+1}, u_t), 
\end{equation}
where we trigger the transition model to predict a next temporal state $d_{t+1}$ to produce a new prior.

The above can be seen as the model's belief over the observable features $f_t$ at the latest time step $t$ under static and change assumptions, respectively. Given a new input $f_{t+1}$, the updated posterior is computed by 
\begin{equation}
q = q_\phi (v_{t+1} | f_{t+1}, d_{t+1}, u_t).
\end{equation}
% where we use $q_{\phi}$ to approximate the true posterior belief state $p_{st}$, here we omit the level index for better readability since we only consider the detection in level $2$. 
% This can be seen as the model's belief over the observable features $f_t$ at the latest time step $t$. 
We then use the KL-divergence, $D_{st}=D_{KL}(q || p_{st})$ and $D_{ch}=D_{KL}(q || p_{ch})$, to measure how much the features have changed compared to the last time step under different assumptions. A change-point boundary is considered to be detected when the static assumption based update is less accurate than the change assumption based update. Particularly, we define such boundary condition as 
\begin{equation}\label{eq:boundary}
    D_{KL}(q||p_{st}) > \beta D_{KL}(q||p_{ch}),
\end{equation}
where $\beta$ is an empirical hyperparameter, $\beta \in [0.15, 0.9]$. Satisfying this criterion indicates that the model’s prediction produced a belief state more consistent with the change assumption, suggesting that it contains a change-point in the features. % that are represented in the second level of the hierarchy. 
In other words, Equ.~\eqref{eq:boundary} compares the difference between the predicted result and observation.

The GRUs, used in the transition model, nevertheless, have been observed to suffer from the state saturation problems in very long sequences~\cite{chang2018temporal,gu2020improving}, and this issue is further aggregated for highly heterogeneous spatial-temporal features as in videos. To counter this problem, we propose to utilize network resetting in our GRU model for $d_t$. That is, after we detect a change-point frame $x_t$, the network is reset to take the next observation $x_{t+1}$ as an initial input, together with the initialization of network parameters. Furthermore, we leverage a dynamic $\beta$ in the boundary condition in Equ.~\eqref{eq:boundary} to preserve stable detection by the following rule:
\begin{equation}\label{eq:dynamic}
    \beta (t+1) = \begin{cases} 
        &\beta (t) + \alpha \textrm{~ (change-point)}; \\
        &\beta (t) - \alpha  \textrm{~ (no-change-point)},
    \end{cases}
\end{equation}
where $\alpha$ is a hyperparameter that is set as $\alpha = 0.15$, and (no-)change-point means that we detect a (no-)change-point at the frame $x_t$.

%At the heart of change-points detection is a mechanism that is used for detecting predictable changes in the observable features over time, and thus determine the boundaries of events. Occurring at the second level of the model, the event detection is used for controlling the structure of the unrolled model over time by allowing (Figure~\ref{fig:3}) or disallowing (Figure~\ref{fig:3}) propagation of the first level to the second level of the hierarchy.

%Action instance is detected with the use of the model’s transition model that computes the next subjective timestep’s belief state, $p_{ch} = p_\theta(s_{t+1}^2|s_t^2,s_{<t}^,s_t^{>n}$. This step can be seen as the model’s prior belief over the features it expects to observe next (the change assumption). Upon receiving the new information $x_{t+1}^n$ , the model then calculates the posterior belief, $q_{ch} = q_\theta(s_{t+1}^n|x_{t+1}^n, s_t^n,s_{<t}^n,s_t^{>n}$. Similarly to the computations for the static assumption, the KL divergence between these two distributions is calculated, $D_{ch}=D_{KL}(q_{ch}||p{ch})$. A change-point boundary is considered detected if $D_{KL}(q_{st}||p{st}) > \beta D_{KL}(q_{ch}||p{ch})$, where $\beta = 0.15$. Satisfying this criterion indicates that the model’s prediction produced a belief state more consistent with the new observation $x_{t+1}^n$, suggesting that it contains a predictable change in the features that are represented in the corresponding level of the hierarchy.

\subsection{Attention-based Classification Module}

The attention-based classification module learns the attention of features for optimizing change-points by distinguishing foreground and background. 

For the classification module, we follow previous work~\cite{qu2021acm,ji2021weakly,shi2020weakly}, applying the cross-entropy loss function between the predicted video-level label and the ground truth label to classify different action classes in a video.
\begin{align}
  \mathcal{L}_{clf}= \sum_{c=1}^{C+1}-y_c(x)\log(p_c(x)),
\end{align}
where $y_c(x)$ is the ground truth video action probability distribution and $p_c(x)$ is the predicted video-level action probability distribution.

For the attention module, we utilize three-branch class activation sequences to represent the foreground, context, and background individually. In specific, we set the video-level instance label
\begin{equation}\label{eq:yn}
    y = (y(1)=0, ..., y(n)=1, ..., y(C)=0, y(C+1)=0),
\end{equation}
which represents the ground truth of the video in the $n$-th category and $C+1$ is the background label index. We then optimize the following attention-based \textit{foreground} model:
\begin{align}
  \mathcal{L}_{fg}= \sum_{c=1}^{C+1}-y(x)\log(p_c(x)).
\end{align}
With the foreground attention weighting, background and action context snippets have been suppressed, as shown in temporal class activation sequences (CAS)~\cite{zhou2016learning,qu2021acm}. 

Similarly, we can set the video-level background label $y = (y(n)=0, y(C+1)=1)$  and context label $y = (y(n)=1, y(C+1) =1)$ to optimize the attention-based \textit{background} model $\mathcal{L}_{bg}$ and \textit{context} model $\mathcal{L}_{ct}$. Following analogical arguments as in the foreground attention branch, we implement contexts and background branches by using CAS.

After obtaining three attention-based classification loss $ \mathcal{L}_{fg}$,  $ \mathcal{L}_{bg}$, and $ \mathcal{L}_{ct}$, we compose the overall loss $\mathcal{L}_{EFC}$ for extraction of foreground change-points as:
\begin{align}
    \mathcal{L}_{EFC}= \mathcal{L}_{fg} + \mathcal{L}_{bg} + \mathcal{L}_{ct}.
\end{align}
For inference, based on the attention-based classification module, we choose the change-points of the foreground as action boundaries. In addition, we add the longest common sub-sequence(LCS) contrasting to optimize the boundaries from the DFC module. In specific, for two adjacent change-points $A$ and $B$, we construct two snippets $l_{AC}=\{l_1, l_2, ...\}$ and $l_{BC}=\{m_1, m_2, ...\}$ by connecting $A$ and $B$ with a third change-point $C$. We then calculate the cosine similarity of $l_i$ and $m_j$ ($i,j = 1,2,...$) and compare the results with a given threshold $0.65$ to get the LCS of $l_{AC}$ and $l_{BC}$. Based on the length of LCS, we delete the redundant points from adjacent change-points.

\section{Experiment}
\label{sec:Experiment}

\setlength{\tabcolsep}{4pt}
\begin{table*}[ht]
\begin{center}
\resizebox{0.95\textwidth}{!}{
\begin{tabular}{|c|c|c|c|c|c|c|c|c|c|c|}
\hline\noalign{\smallskip}
\multicolumn{1}{|c|}{{Type}} & {Model} & {Publication} & \multicolumn{8}{|c|}{THUMOS14}  \\

\multicolumn{1}{|c|}{}& & &0.1& 0.2 &0.3 & 0.4 & 0.5  & 0.6 & 0.7 &avg. \\
\noalign{\smallskip}
\hline
\noalign{\smallskip}

{}& BU-TAL~\cite{zhao2020bottom} & \textit{ECCV20} & - & - & 53.9 & 50.7 & 45.4  &38.5 &28.0 & - \\

{}& G-TAD~\cite{xu2020g} & \textit{CVPR20} &-& - & 54.5 & 47.6 & 40.2  & 30.8 & 23.4 & - \\

{Fully-supervised}& GCM~\cite{zeng2021graph} &  \textit{TPAMI21} &72.5& 70.9 & 66.5 & 60.8 & 51.9  & - & - & - \\

{}& AFSD~\cite{lin2021learning} &  \textit{CVPR21} &72.2& 70.8 & 67.1 & 62.2 & 55.5 & 43.7 & 31.1 & 57.6\\

{}& TadTR~\cite{liu2021end} &  \textit{TIP22} & - & - & 74.8 & 69.1 & 60.1 & 46.6 & 32.8 & - \\

{}& RefactorNet~\cite{liu2021end} &  \textit{CVPR22} & - & - & 70.7 & 65.4 & 58.6 & 47.0 & 32.1 & - \\

{}& TRA~\cite{zhao2022temporal} &  \textit{TIP22} & 73.7 & 72.6 & 70.0 & 64.3 & 57.4 & 46.2 & 31.1 & 59.3 \\
\hline
{}& CMCS~\cite{liu2019completeness} &  \textit{CVPR19} &57.4& 50.8 & 41.2 & 32.1  & 23.1 &15.0 &7.0 &32.4 \\

{}& WSAL-BM~\cite{nguyen2019weakly} &  \textit{ICCV19} &60.4& 56.0 & 46.6 & 37.5 & 26.8  & 19.6 &9.0 &36.6 \\

{}& DGAM~\cite{shi2020weakly} &  \textit{CVPR20} &60.0& 54.2 & 46.8 & 38.2 & 28.8  & 19.8 &11.4 &37.0  \\

{}& A2CL-PT~\cite{min2020adversarial} &  \textit{ECCV20} &61.2 &56.1 &48.1 &39.0 &30.1 &19.2 &10.6 &37.8 \\

{}& HAM-Net~\cite{islam2021hybrid} &  \textit{AAAI21} &65.9 &59.6 &52.2 &43.1 &32.6 &21.9 &12.5 &41.1\\ 

{Weakly-supervised}& FAC-Net~\cite{huang2021foreground} &  \textit{ICCV21} &67.6 &62.1 &52.6 &44.3 &33.4 &22.5 &12.7 &42.2 \\ 

{}& CoLA~\cite{zhang2021cola} &  \textit{CVPR21} &66.2 &59.5 &51.5 &41.9 &32.2 &22.0 &13.1  &40.9 \\ 

{}& ACM-Net~\cite{qu2021acm} &  \textit{TIP21} &68.9 &62.7 &55.0 &44.6 &34.6 &21.8 &10.8 &42.6\\ 

{}& FTCL~\cite{gao2022fine} &  \textit{CVPR22} &69.6 &63.4 &55.2 &45.2 &35.6 &23.7 &12.2 &43.6\\ 

{}& ASM-Loc~\cite{he2022asm} &  \textit{CVPR22} &71.2 &65.5 &57.1 &46.8&36.6 &25.2 &13.4 &45.1\\ 
{}& DCC~\cite{li2022exploring} &\textit{CVPR22}& 69.0 & 63.8 & 55.9 & 45.9 & 35.7 & 24.3 & 13.7 & 44.0\\
{}& RSKP~\cite{huang2022weakly} &\textit{CVPR22}& 71.3 &  65.3& 55.8 &47.5 &38.2& 25.4 &12.5& 45.1\\
{}& DELU~\cite{chen2022dual} &\textit{ECCV22}& 71.5 &66.2 &56.5& 47.7& \textbf{40.5} &\textbf{27.2} &\textbf{15.3}& 46.4\\
{}& StochasticFormer~\cite{shi2023Stochast} &  \textit{TIP23} &66.5 &61.1 &52.5 &43.9 &33.5 &22.6 &13.2 &41.9\\ 
{}& ASCN~\cite{shi2023Stochast} &  \textit{TMM23} &71.4 &65.6 &57.0 &48.2 &39.8 &26.8 &14.4 &46.1\\ 
{}& Ours & - &\textbf{75.1}& \textbf{68.9}& \textbf{60.2} & \textbf{48.9} & 38.3  & 26.8 & 14.7 & \textbf{47.2}\\
\hline
\end{tabular}}
\end{center}
\caption{
% Performance comparison with state-of-the-art methods on THUMOS14 and ActivityNet v1.3, measured by mAP at different IoU thresholds on THUMOS14 and ActivityNet v1.3.
Performance comparison with SotA methods on THUMOS14, measured by mAP at different IoU thresholds.
}
\label{table:1}
\end{table*}
\setlength{\tabcolsep}{4pt}
In this section, we first describe datasets and evaluation metrics. Then, we evaluate our model's effectiveness followed by the main results and ablation study.

\subsection{Datasets and Evaluation Metrics}
To validate the effectiveness of our model, we conduct extensive experiments on commonly-used benchmark THUMOS14~\cite{THUMOS14} and ActivityNet v1.3~\cite{caba2015activitynet}. 

\textbf{THUMOS14~\cite{THUMOS14}} It contains $101$ categories of videos and is composed of four parts: training, validation, testing and a background set. Each set includes $13320$, $1010$, $1574$ and $2500$ videos, respectively. Following the common setting in~\cite{THUMOS14}, we used $200$ videos in the validation set for training, and $213$ videos in the testing set for evaluation.

\textbf{ActivityNet~\cite{caba2015activitynet}} We evaluate our method on the ActivityNet release 1.3, which contains samples from $200$ categories of activities and $19994$ videos in total. It includes untrimmed video classification and activity detection tasks. It is divided into training, validation, and test sets with a ratio of $2:1:1$, containing $10024$, $4926$ and $5044$ videos respectively. Following~\cite{he2022asm,qu2021acm}, we use the training set to train our model and the validation set for evaluation.

\textbf{Evaluation Metrics} We follow the standard evaluation protocol and report mean Average Precision (mAP) at the different intersections over union (IoU) thresholds. The results are calculated using the benchmark code provided by the ActivityNet official codebase\footnote{https://github.com/activitynet/ActivityNet/tree/master/Evaluation}.

\subsection{Implementation Details}

For the feature extraction, we first sample RGB frames at 25 fps for each video and apply the TV-L1 algorithm~\cite{zeng2019graph} to generate optical flow frames. Then, we divide each video into non-overlapping snippets with consecutive 16 frames. Thereafter, we perform the I3D
networks~\cite{carreira2017quo} pre-trained on the Kinetics dataset~\cite{carreira2017quo} to obtain the video features. Note that, for a fair comparison, we do not introduce any other feature fine-tuning operations to the pre-trained I3D model.

For the training process on the THUMOS-14, we set the batch size as $16$, and apply the Adam optimizer~\cite{kingma2014adam} with learning rate $10^{-4}$ and weight decay $5 \times 10^{-4}$. We set the video snippets length $T=750$. For the training process on the ActivityNet-v1.3 dataset, we set the training video batch size to $64$, applying the same Adam optimizer with THUMOS-14, and set the video snippets length $T=150$.

\subsection{Main Results}
In Table~\ref{table:1}, we compare AHLM with current methods on Thumos14. Selected fully-supervised methods are presented for reference. We observe that AHLM outperforms all the previous WSAL methods and establishes new state of the art on THUMOS-14 with $47.2\%$ average mAP for IoU thresholds $0.1:0.7$. In particular, our approach outperforms ACM~\cite{qu2021acm}, which also utilizes an attention model to distinguish foreground and background but without explicit temporal variations of feature semantics. Even compared with the fully supervised methods, AHLM outperforms BU-TAL and GTAD and achieves comparable results with AFSD and TRA when the IoU threshold is low. The results demonstrate the superior performance of our approach with temporal variations of feature semantics modeling.

We also conduct experiments on ActivityNet-v1.3 and the comparison results are summarized in Table~\ref{table:2}. Again, our AHLM obtains a new state-of-the-art performance of $25.9\%$ average mAP, surpassing the latest works (e.g. FTCL, ASM-Loc). The consistent superior results on both datasets justify the effectiveness of our AHLM. Especially, unlike other methods which only achieve marginal improvement on ActivityNet-v1.3, AHLM maintains the significant improvement as on THUMOS-14. This shows that the performance of AHLM does not rely on the length of video snippets, since the average length of videos in the THUMOS14  dataset is much longer than those in ActivityNet-v1.3. An important reason is that AHLM learns the temporal variations of the feature semantics in the latent space by GRU with dynamic resetting, which is appropriate to detect temporal variations of semantics in different timescales.

% Specially, for high IoU set, the performance of ActivityNet-v1.3 is better than THUMOS-14. In the experiment implementation, we set the ActivityNet-v1.3 video snippet length shorter than THUMOS-14. The videos in THUMOS14 dataset contain a large proportion of background information, and each video contains more than one action instance from one or multiple classes, but each video of ActivityNet v1.3 contains $1.65$ action instances from one or multiple classes on average. An important reason is that AHLM learns the temporal variations of the feature semantics in the latent space by GRU, which is more appropriate for short-term temporal semantics and less background.
\begin{table}[ht]
  \begin{center}

  \resizebox{0.45\textwidth}{!}{
  \begin{tabular}{|c|c|c|c|c|c|}
  \hline\noalign{\smallskip}
  \multicolumn{1}{|c|}{Model}& {Publication} &\multicolumn{4}{|c|}{ActivityNet v1.3}  \\
  \cline{3-6}
  \multicolumn{1}{|c|}{}&{}&0.5& 0.75 &0.95 &avg. \\
  \hline
  STPN~\cite{nguyen2018weakly} &  \textit{CVPR 2018} &29.3 &16.9 &2.6 &16.3\\
  ASSG~\cite{zhang2019adversarial} &  \textit{MM 2019} &32.3 &20.1 &4.0 &18.8\\
  Bas-Net~\cite{lee2020background} &  \textit{AAAI 2020} &34.5 &22.5 &4.9 &22.2\\
  TS-PCA~\cite{yang2021uncertainty} &  \textit{CVPR 2021} &37.4 &23.5 &5.9 &23.7\\
  FAC-Net~\cite{huang2021foreground} & \textit{ICCV 2021} &37.6 &24.2 &6.0 &24.0\\
  ACM-Net~\cite{qu2021acm} & \textit{TIP 2021} &37.6 &24.7 &6.5 &24.4\\
  FTCL~\cite{gao2022fine} & \textit{CVPR 2022} &40.0 &24.3 &6.4 &24.8\\
  ASM-Loc~\cite{he2022asm} & \textit{CVPR 2022} &41.0 &\textbf{24.9} &6.2 &25.1\\
  Ours & - & \textbf{42.3}& 24.8 & \textbf{6.9} & \textbf{25.9}\\
  
  \hline
  \end{tabular}}
  \end{center}
   \caption{
  % Performance comparison with state-of-the-art methods on THUMOS14 and ActivityNet v1.3, measured by mAP at different IoU thresholds on THUMOS14 and ActivityNet v1.3.
  Performance comparison with state-of-the-art methods on ActivityNet-v1.3 dataset. }
  \label{table:2}
  \end{table}
\subsection{Ablation Study}

To demonstrate the reasonableness of our AHLM, we analyze the effect of every submodule and some function operations in this subsection.
\subsubsection{Contribution of each design in AHLM}
We study the influence of each component in AHLM on overall performance. We start with the basic model that directly optimizes the attention based on foreground loss $ \mathcal{L}_{fg} $. Then we add the background loss $ \mathcal{L}_{bg} $ and the context loss $ \mathcal{L}_{ct} $ gradually. These three types of loss constitute $\mathcal{L}_{EFC}$. The variational generative model loss $ \mathcal{L}_{ELBO} $, which indicates $\mathcal{L}_{DFC}$, is further included. Note that adding $\mathcal{L}_{ELBO}$ indicates involving the hierarchically-structured modeling, where reconstruction loss of VAE is also simultaneously optimized. 

Table~\ref{table:3} summarizes the performances by considering one more factor at each stage on THUMOS14. We first observe that adding the background loss $ \mathcal{L}_{bg} $ and the context loss $L_{ct}$ largely enhances the performance of the foreground-based model. The two losses encourage the sparsity in the foreground attention weights by pushing the background attention weights to be $1$ at background snippets, and therefore improve the foreground-background separation. Based on the EFC, our change-point detection module further contributes a significant increase of $2.4\%$ and the performance of AHLM finally reaches $47.0\%$. Further, a more explicit ablation study by adding DFC on each part of the EFC proves our method's effectiveness, in particular, our DFC module contributes an increase of $2.1\%$ and $2.5\%$ respectively based on foreground and foreground-background.
\begin{table}[ht]
  \begin{center}
  \resizebox{0.45\textwidth}{!}{
    \begin{tabular}{|c c c|c|c|c|c|c|c|}
      \hline
      \multicolumn{3}{|c|}{$\mathcal{L}_{EFC}$}&\multicolumn{1}{|c|}{$ \mathcal{L}_{DFC} $}&\multicolumn{5}{|c|}{THUMOS14} \\
      \cline{1-9}
      \multicolumn{1}{|c}{{$ \mathcal{L}_{fg} $} }&\multicolumn{1}{c}{ {$ \mathcal{L}_{bg} $} }  &\multicolumn{1}{c|}{ {$ \mathcal{L}_{ct} $} }&\multicolumn{1}{|c|}{$ \mathcal{L}_{ELBO} $}&0.1 &0.3 & 0.5 & 0.7 &avg.\\
      \hline
      \checkmark &- &-  &- &49.9 &32.9 &16.6 &5.3 &26.2\\
      \checkmark &- &-  &\checkmark &\textbf{54.4} &\textbf{36.4} &\textbf{16.8} &\textbf{5.4} &\textbf{28.3}\\
      \checkmark & \checkmark &- &- &55.9 &41.9 &23.0 &7.1 &32.0\\
      \checkmark & \checkmark & - &\checkmark  &\textbf{57.3} &\textbf{44.2} &\textbf{26.6} &\textbf{10.3} &\textbf{34.5} \\
      \checkmark & \checkmark & \checkmark &-  &71.2 &57.1 &36.6 &13.4 &44.6 \\
      \checkmark & \checkmark & \checkmark & \checkmark  &\textbf{75.1} &\textbf{60.2}  &\textbf{38.3} &\textbf{14.7} &\textbf{47.0}\\
      \hline
      \end{tabular}}
  \end{center}
\caption{\textit{m}AP at different overlap IoU thresholds performance comparison of each design on THUMOS14.}
\label{table:3}
\end{table}
\subsubsection{Effectiveness of Change-point Module}
It is obvious that the proposed change-point detection strategy can play a complementary role over existing methods in localizing boundaries of action instances. To see this, we conduct the experiment by directly adding the detected points by our change-point detection module into a MIL-based method~\cite{paul2018w}, which indicates the classification method with CAS~\cite{zhou2016learning}. Specifically, for the proposed snippets based on the change-points, we calculate the score using the MIL-based method. 

Table~\ref{table:4} shows, compared to the original MIL-based method, our change-point detection module contributes a significant increase of $3.7\%$  and the performance finally reaches $28.5\%$ on average map.
% In Table~\ref{table:4}, we evaluate the impact of change-point module for localizing action boundaries. Based on the change-point module, we add the classification module for evaluating WSAL task. The MIL-based method~\cite{paul2018w} indicates classification method with CAS~\cite{zhou2016learning}, which does not include the attention model. Our change-point detection module also contributes a significant increase of $8.7\%$  and the performance finally reaches $28.5\%$ on average map.
\begin{table}[ht]
  \begin{center}

  \resizebox{0.45\textwidth}{!}{
    \begin{tabular}{|c|c|c|c|c|c|}
      \hline
      \multicolumn{1}{|c}{}& \multicolumn{5}{|c|}{THUMOS14} \\
      \cline{2-6}
      \multicolumn{1}{|c|}{}&0.1 &0.3 & 0.5 & 0.7 & avg.\\
      \hline
      MIL-based~\cite{paul2018w} &46.5 &31.2 &16.9 &4.4 &24.8\\
      % \cline{1-8}
      MIL-based + Ours &\textbf{54.2} & \textbf{37.1} &\textbf{17.3}  &\textbf{5.3}  &\textbf{28.5} \\
      \hline
      \end{tabular}}
  \end{center}
\caption{Comparison with MIL-based methods on THUMOS14.}
\label{table:4}
\end{table}

Following the common setting of WSAL task, as Table~\ref{table:1} and Table~\ref{table:2} show, we
chose THUMOS14 and ActivityNet dataset as our benchmark and achieves SOTA performance. Regarding the scalability and generalization, essentially, 
the effectiveness of our change-point module is related to the representation ability of the hierarchical-VAE, which is proved in the 
literature(\textit{e.g.}, CW-VAE~\cite{saxena2021clockwork}, NVAE~\cite{vahdat2020nvae}, VPR~\cite{zakharov2021variational}), hence guarantees the scalability of our change-point module.
\begin{figure}[ht]
  \centering
  \includegraphics[width=0.46\textwidth]{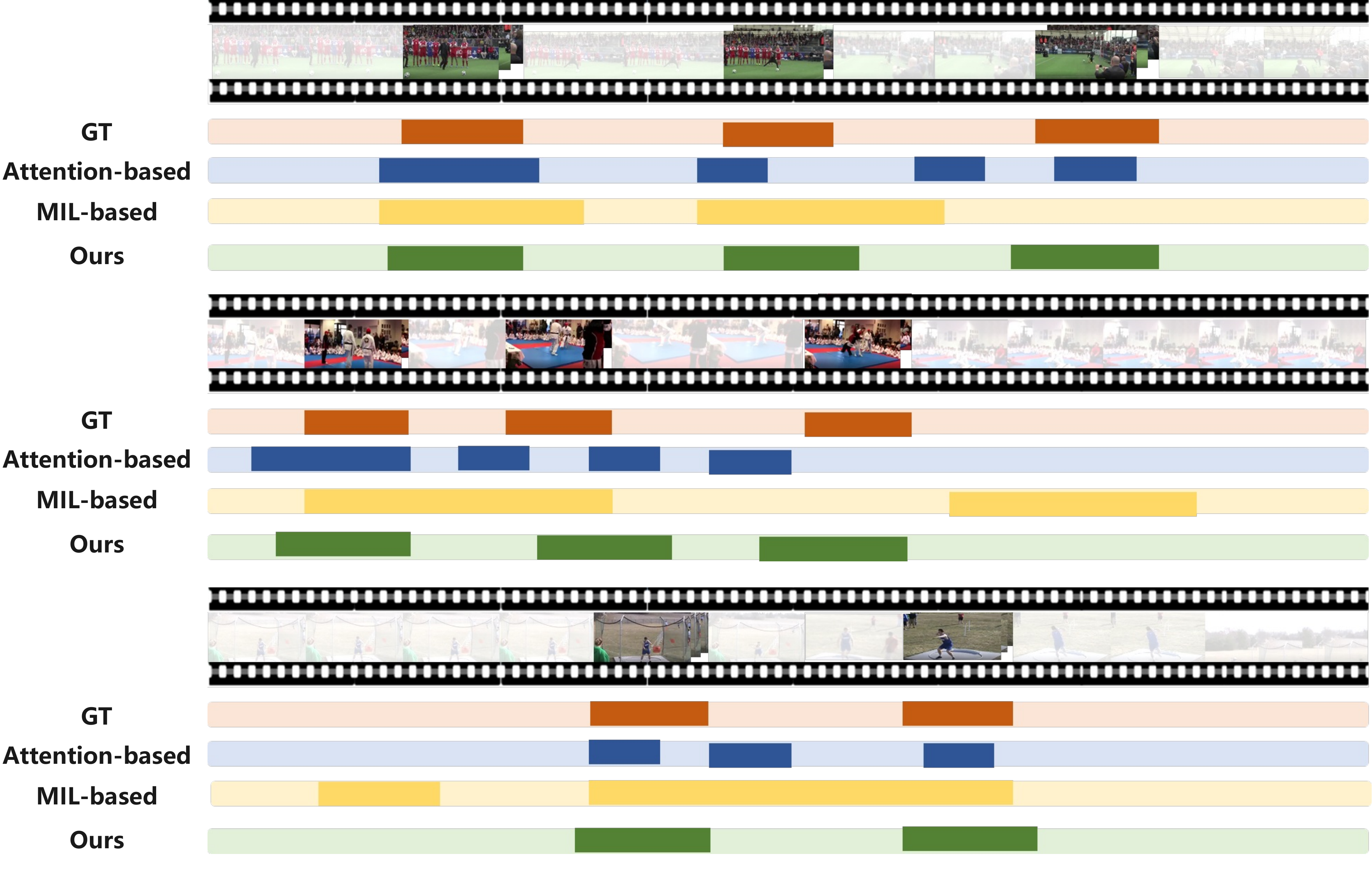}
  \caption{Qualitative result visualization on the THUMOS-14 dataset. From the above qualitative results, we can conclude that our proposed AHLM mechanism is greatly beneficial to suppress locate action instances and help us achieve more precise temporal action localization results.}
  \label{fig:4}
\end{figure}
\subsubsection{Qualitative Results}

Fig \ref{fig:4} shows the visualization comparisons between the attention-based model~\cite{qu2021acm}, MIL-based model~\cite{paul2018w} and our AHLM. From the figure, it can be found that the common errors of current methods are mainly about missed detection of short actions and imprecise localization of the action. Through learning temporal variation of feature semantics, AHLM locates the boundaries of short actions (\textit{e.g.}, examples $1$) and more precise boundaries (\textit{e.g.}, examples $2, 3$). We see our method depends on an attention-based classification model to filter the change-points from fore- and background. This verifies the importance of improving the quality of separating foreground and background and should be further studied in future work.

\section{Conclusion}
\label{sec:Conclusion}

In this paper, we propose a novel hierarchically-structured attention mechanism to model temporal variation of feature semantics by disentangling the spatial and temporal information on the latent space. Our weakly supervised action localization framework AHLM mainly consists of feature embedding, change-point detection module and attention-based classification module. We leverage temporal variation of the features to locate the change-points and optimize by attention-based classification model for the WSAL task. Our method outperforms the prior work with a remarkable margin on two popular datasets, achieving the SotA results on both.
% on THUMOS14, achieving the SotA result (around $46.0\%$ average mAP), as well as on ActivityNet v1.3 (around $25.3\%$ average mAP). 
The results demonstrate that our exploration on temporal variation of feature semantic information effectively improves WSAL performance, which narrows the performance gap between the weakly-supervised and fully-supervised settings. For the future work, we believe the hierarchically-structured latent modeling will be a promising direction for various weakly supervised and unsupervised learning tasks. It is also worth to explore such mechanism in other related tasks.

\textbf{Acknowledgement.} we want to thank Alexey Zakharov and Zafeirios Fountas from Huawei UK, for many helpful discussions and inspirations. 

{\small
\bibliographystyle{ieee_fullname}
\bibliography{egbib}
}

\end{document}